\documentclass[journal]{IEEEtran}
\usepackage{graphicx,cite,amsmath,amssymb,autobreak,hhline,booktabs,stfloats,bm,amsthm,hyperref,times,subfigure,multirow,algpseudocode,}
\usepackage{algorithm,algpseudocode,multicol,threeparttable,dcolumn,latexsym,rotating,textcomp,colortbl,enumerate,tabularx}
\usepackage{array}         % 提供表格列宽和对齐方式的支持（如 p{} 和 \raggedright）
\usepackage{graphicx}      % 支持自动换行（用于增强 \raggedright 效果）
\usepackage{booktabs}      % 可选，提供更专业的表格线条样式（如 \toprule、\midrule 等）
\usepackage{caption}       % 用于调整标题样式

\usepackage{verbatim}
\usepackage{pifont} % For the star symbols
\usepackage{array}  % For flexible column widths
\usepackage{graphicx} % To adjust column content if needed

\usepackage{verbatim}
\usepackage{caption}
\usepackage[top=2cm, bottom=2cm, left=2cm,right=2cm]{geometry} 
\usepackage{graphicx}
\usepackage{grffile}
\usepackage{stfloats}
\usepackage{dsfont}
\allowdisplaybreaks[4]

\begin{document} 

\newtheorem{lemma}{\bf Lemma }
\newtheorem{theorem}{\bf Theorem}	
\newtheorem{corollary}{Corollary}
\renewcommand{\algorithmicrequire}{\textbf{Input:}} 
\renewcommand{\algorithmicensure}{\textbf{Output:}}
\newtheorem{remark}{Remark}
\newtheorem{proposition}{\bf Proposition }

\title{Federated Fine-Tuning of LLMs: Framework Comparison and Research Directions}

% Unlocking Collaborative Fine-Tuning of LLMs: A Comparative Analysis of Federated Frameworks
%Unlocking Collaborative Fine-tuning of LLMs: Knowledge Sharing in Federated Approach

%Synergy Between Federated Learning and Large Language Models: A Collaborative Approach
 \author{
    Na Yan, Yang Su, Yansha Deng, \IEEEmembership{Senior Member,~IEEE}, and Robert Schober, \IEEEmembership{Fellow,~IEEE}
    \thanks{
        Na Yan, Yang Su, and Yansha Deng are with the Department of Engineering, King's College London, London, WC2R 2LS, U.K. (e-mails: \{na.2.yan, yang.2.su, yansha.deng\}@kcl.ac.uk).
        
        Robert Schober is with the Institute for Digital Communication, Friedrich-Alexander-Universit\"at Erlangen-N\"urnberg, 91052 Erlangen, Germany (e-mail: robert.schober@fau.de).
    }
}

\maketitle
\begin{abstract}
Federated learning (FL) provides a privacy-preserving solution for fine-tuning pre-trained large language models (LLMs) using distributed private datasets, enabling task-specific adaptation while preserving data privacy. 
However, fine-tuning the extensive parameters in LLMs is particularly challenging in resource-constrained federated scenarios due to the significant communication and computational costs.
To gain a deeper understanding of how these challenges can be addressed, this article conducts a comparative analysis three advanced federated LLM (FedLLM) frameworks that integrate knowledge distillation (KD) and split learning (SL) to mitigate these issues: 1) FedLLMs, where clients upload model parameters or gradients to enable straightforward and effective fine-tuning; 2) KD-FedLLMs, which leverage KD for efficient knowledge sharing via logits; and 3) Split-FedLLMs, which split the LLMs into two parts, with one part executed on the client and the other one on the server, to balance the computational load.
Each framework is evaluated based on key performance metrics, including model accuracy, communication overhead, and client-side computational load, offering insights into their effectiveness for various federated fine-tuning scenarios.
Through this analysis, we identify framework-specific optimization opportunities to enhance the efficiency of FedLLMs and discuss broader research directions, highlighting open opportunities to better adapt FedLLMs for real-world applications.
A use case is presented to demonstrate the performance comparison of these three frameworks under varying configurations and settings.


\end{abstract}

\section{Introduction}
\begin{figure*}
    \centering
    \includegraphics[width=1\linewidth]{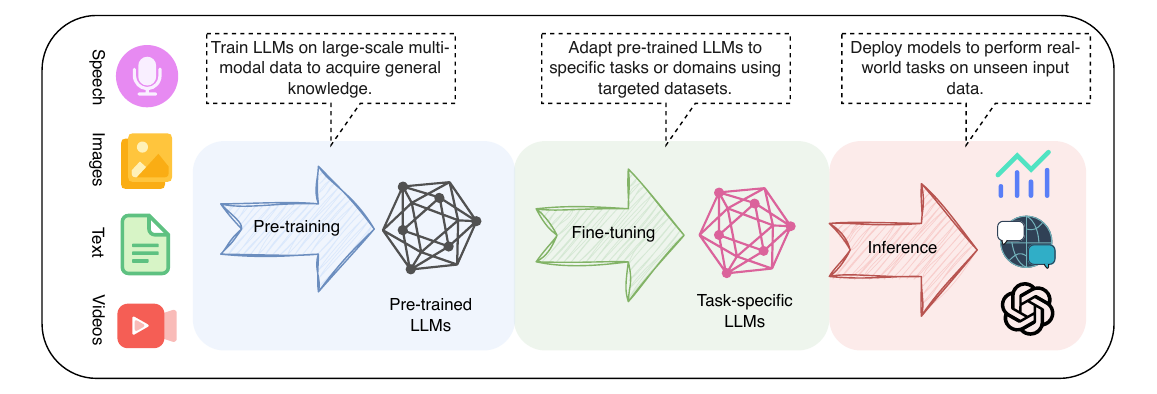}
    \caption{Three life stage of LLMs: Pre-training, fine-tuning, inference.}
    \label{fig1}
\end{figure*}

Large language models (LLMs) represent a significant advancement in artificial intelligence, achieving exceptional performance in tasks such as natural language understanding and generation \cite{chang2024survey}. 
These models have found applications across diverse domains and tasks, including chatbots (e.g., ChatGPT), writing assistants (e.g., LanguageTool), and advanced search engines (e.g., New Bing). 
Before these task-specific LLMs can be deployed for inference services, they typically undergo two training stages, as illustrated in Fig. \ref{fig1}. 
During pre-training, LLMs are trained on large-scale public datasets to acquire a broad understanding of language. In the fine-tuning stage, the models are adapted to specialized tasks using task-specific data. However, centralizing these distributed private task-specific data for fine-tuning presents significant challenges, particularly due to privacy concerns and regulatory constraints on data access.

Federated learning (FL) facilitates the privacy-preserving fine-tuning of pre-trained LLMs on distributed clients by only sharing model updates between the server and clients, ensuring the data remains localized \cite{hilmkil2021scaling}.
However, fine-tuning large models like GPT-3 (175 billion parameters) \cite{brown2020language}, BERT (340 million parameters) \cite{kenton2019bert}, and LLaMA (65 billion parameters) \cite{zhuang2023foundation} in resource-constrained federated scenarios presents significant challenges, including high communication and computational overhead.

To address these challenges, parameter-efficient fine-tuning (PEFT) \cite{ding2023parameter} has  been proposed, focusing on fine-tuning only a subset of the model's parameters. Building upon the fundamental FedLLM framework using PEFT, knowledge distillation (KD) and split learning (SL) are integrated to further enhance fine-tuning efficiency. These combined approaches have led to to the development of three advanced FedLLMs fine-tuning frameworks.
The first framework, FedLLMs, is the foundational approach, where clients directly transmit model updates to the server. The second framework, KD-FedLLMs, employs KD to enable efficient knowledge transfer through logits exchanged between the global and local models \cite{fan2024fedmkt}. The third framework, Split-FedLLMs, integrates SL with FedLLMs, facilitating the exchange of intermediate activations between clients and the server \cite{lin2024splitlora}. Each framework highlights a distinct approach for knowledge transfer, showcasing their unique strategies for federated fine-tuning.

%Each of these frameworks demonstrates distinct performance characteristics in the FedLLMs fine-tuning process, offering unique trade-offs in terms of communication, computation, and model accuracy. 

To evaluate the performance of these frameworks, including the effectiveness of fine-tuning, as well as their communication and computational overhead, this article offers a systematic overview and comparison of these three FedLLM fine-tuning frameworks, while identifying key optimization opportunities.
To the best of our knowledge, this is the first comprehensive comparison of these FedLLMs frameworks. The main contributions of this article are:
\begin{itemize}
    \item \textbf{Comparison of Three FedLLM Frameworks}: This article presents a comprehensive overview of three FedLLM frameworks: basic FedLLMs, KD-FedLLMs, and Split-FedLLMs. By examining the distinct knowledge transfer methods based on direct model updates, knowledge sharing via logits, and intermediate activations, this work provides valuable insights into how each framework addresses the efficiency challenges of federated fine-tuning.

\item \textbf{Comprehensive Evaluation of Key Metrics}: The article evaluates these frameworks based on key performance metrics, including model accuracy, communication overhead, and computational load. This evaluation helps in understanding the trade-offs between different methods and provides practical insights into their suitability for diverse federated fine-tuning scenarios.

\item \textbf{Framework-specific Optimization Opportunities and Broader Research Directions}: By analyzing the factors impacting each framework, this work identifies framework-specific optimization opportunities to improve their efficiency. We further explore open research directions, such as continual learning for FedLLMs and multi-modal support, to better adapt FedLLMs to real-world applications.

\item \textbf{Practical Use Case Demonstration}: This article concludes with a use case that exemplifies the practical application of these three frameworks, offering real-world insights into their performance and effectiveness under various settings.
\end{itemize}

\section{Federated Frameworks for LLM Fine-tuning}
This section introduces the principles of the three FedLLM frameworks \footnote{All these frameworks fine-tune LLMs utilizing PEFT, as it offers a practical and scalable solution for model adaptation in resource-constrained federated scenarios due to its reduced resource requirements. 
These frameworks are also amenable to full parameter fine-tuning approaches.}. 
\begin{figure*}
    \centering
    \includegraphics[width=1\linewidth]{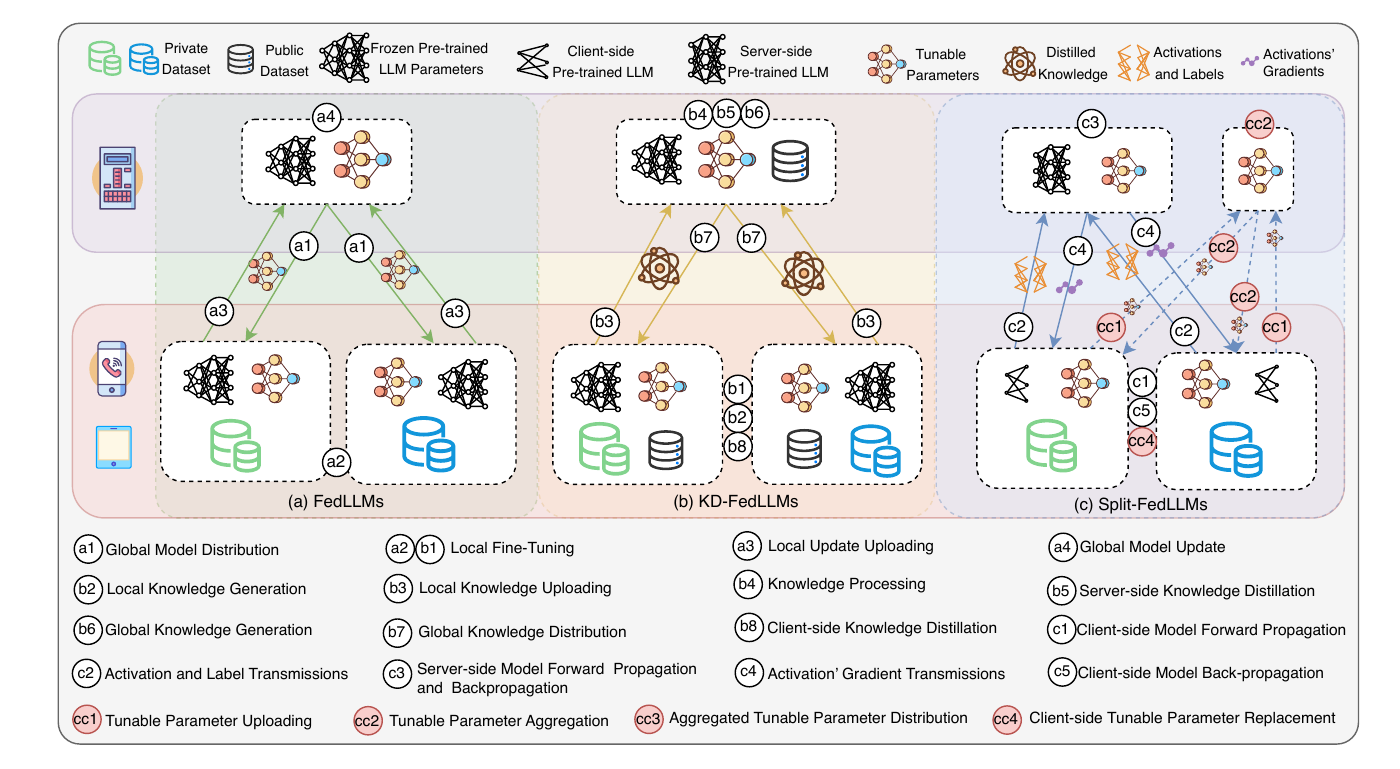}
    \caption{Overview of three federated fine-tuning frameworks: (a) basic FedLLMs; (b) KD-FedLLMs; (c) Split-FedLLMs.}
    \label{fig2}
\end{figure*}
\subsection{FedLLMs: Foundational Approach to Federated Fine-Tuning}
FedLLMs, as presented in Fig.~\ref{fig2}(a), represent the foundational approach to federated fine-tuning. 
The system comprises a central server and multiple distributed clients. Each client starts with a frozen pre-trained LLM and fine-tunes a set of tunable parameters, referred to as ``tunable parameters" throughout this paper. These parameters vary based on the specific PEFT technique used, such as Low-Rank Adaptation (LoRA), adapters, or prompts, and are optimized using the client's private dataset.
The central server manages the training process by collecting fine-tuned parameter updates from all clients and aggregating them to refine the global model parameters. The typical workflow for a single training round consists of the following steps:
\begin{itemize} 
\item \textbf{Global Model Distribution and Local Fine-Tuning (a1, a2):} The server sends the global tunable parameters to the participating clients (a1), who then fine-tune the tunable parameters locally using their private datasets (a2). 
\item \textbf{Local Update Uploading and Global Model Update (a3, a4):} Clients upload their fine-tuned tunable parameters to the server, which aggregates them to refine the global tunable parameters for the next iteration. 
\end{itemize}
This iterative process continues until the global model achieves the desired performance, effectively enabling collaborative fine-tuning while ensuring that client data remains private.

\subsection{KD-FedPEFT: Logit-Based Knowledge Sharing} \label{section2-B}
KD is a machine learning technique that enables the transfer of knowledge from a ``teacher" model to a ``student" model \cite{wu2022communication} via logits sharing. The logits are the predicted probabilities of different potential outputs from the ``teacher" model for a given input, guiding the student model to mimic the performance of the ``teacher" model. As shown in Fig. \ref{fig2}(b), the workflow of KD-FedLLMs includes the following main steps:
%For instance, in a classification task with three classes, rather than producing a single class label, the teacher may output probabilities such as 70\% for “cat,” 20\% for “dog,” and 10\% for “rabbit.” 

\begin{itemize}
    \item \textbf{Local Fine-tuning and Client-Side Knowledge Generation and Uploading (b1, b2, b3)}: After local fine-tuning (b1), clients use the fine-tuned local models to generate knowledge representations (logits) of samples in a public dataset (b2) and then upload these representations to the server (b3).
\item \textbf{Knowledge Processing and Distillation (b4, b5)}: The server processes the received knowledge to get a refined global knowledge set (b4), and performs KD to update the global model (b5). This ensures the integration of diverse client learned knowledge into the global model.
\item \textbf{Global Knowledge Generation, Distribution, and Client-side KD (b6, b7, b8)}: The server generates the logits through the refined global model (b6) and distributes them to clients (b7). Each client uses this knowledge to align its local model through KD, improving the local model from the global knowledge (b8).
\end{itemize}
The shared knowledge representations can take other forms, such as feature embeddings from intermediate layers, and softened label distributions, depending on the design of the framework.

\subsection{Split-FedPEFT: Activation-Based Updates}
SL is a collaborative learning paradigm where the model is split between clients and the central server \cite{thapa2022splitfed}. In this approach, the clients process the training data samples and compute the intermediate activations, which are then sent to the server. The server processes these activations and completes the remaining part of the model training.
Split-FedLLMs, as shown in Fig. \ref{fig2}(c) \cite{lin2024splitlora}, apply this approach to LLMs, with the following workflow:
\begin{itemize}
\item \textbf{Client-side Model Forward Propagation and Activation Transmission (c1, c2)}: Clients compute activations using the initial model layers on their private data and then send these activations and labels to the server.
\item \textbf{Server-side Model Forward Propagation and Backpropagation, and Gradient Transmission (c3, c4)}: The server processes activations, computes the loss, and performs backpropagation (i.e., fine-tune the server-side LLMs) (c3). It then sends the gradients for the activations back to the clients (c4).
\item \textbf{Client-side Model Backpropagation and Tunable Parameter Updates (c5, cc1)}: Clients use the received gradients to update their initial layers (i.e., fine-tune the client-side LLMs) (c5), and then upload the fine-tuned tunable parameters to the server (cc1).
\item \textbf{Tunable Parameters Aggregation, Distribution, and Replacement (cc2, cc3, cc4)}: The server aggregates the tunable parameters from all clients (cc2), redistributes them (cc3), and clients replace their local parameters with the aggregated ones (cc4).
\end{itemize}
The LLMs can be split between client and server based on the design of the FedLLM system. A common approach is inter-transformer splitting, where initial transformer blocks are processed by the client, and deeper, more computationally intensive blocks are handled by the server. Alternatively, intra-transformer splitting divides tasks within a single transformer block, such as processing self-attention on the client and the feed-forward network on the server, allowing more flexible resource allocation.

%The choice between inter-transformer and intra-transformer splitting depends on factors like model size, client-server computational capabilities, and the need to optimize communication and efficiency. Both aim to minimize overhead and cost in resource-constrained federated settings while maintaining model performance.

\section{Comparative Analysis and Comparison}\label{section4}
This section quatitively analyzes and compares the three considered frameworks, focusing on key performance metrics, including model accuracy, communication efficiency, and client-side computational load. 
\begin{table*}[h]
\centering
\caption{Comparative analysis of key performance metrics across FedLLM Frameworks.}
\renewcommand{\arraystretch}{1.5} % Adjust row height
\begin{tabularx}{\textwidth}{|>{\centering\arraybackslash}m{3.5cm}|>{\centering\arraybackslash}X|>{\centering\arraybackslash}X|>{\centering\arraybackslash}X|}
\hline
\textbf{Metric} & \textbf{FedLLMs} & \textbf{KD-FedLLMs} & \textbf{Split-FedLLMs} \\ \hline

\textbf{Model Accuracy} & 
\ding{72}\ding{72}\ding{72}\ding{72}\ding{72} & 
\ding{72}\ding{72}\ding{72} & 
\ding{72}\ding{72}\ding{72}\ding{72}\\ \hline

\multirow{2}{*}{\textbf{Communication Overhead}} & 
\multirow{2}{*}{\ding{72}\ding{72}\ding{72}} & 
\ding{72}\ding{72} \textbf{(Classification Task)} & 
\multirow{2}{*}{\ding{72}\ding{72}\ding{72}\ding{72}} \\ \cline{3-3}
& & \ding{72}\ding{72}\ding{72}\ding{72}\ding{72} \textbf{(Generative Task)} & \\ \hline

\textbf{Client-Side Computation Overhead} & 
\ding{72}\ding{72}\ding{72}\ding{72} & 
\ding{72}\ding{72}\ding{72}\ding{72}\ding{72} & 
\ding{72}\ding{72}\ding{72} \\ \hline

\end{tabularx}
\label{table1}
\end{table*}

\subsection{Model Accuracy}
FedLLMs are expected to deliver the best model performance because they directly update the parameters, allowing for more accurate adaptation to the local data on each client without relying on any intermediate steps, as done in the other frameworks.

In Split-FedLLMs, performance may experience slight degradation due to the LLMs being split between client and server. This division limits the client to fine-tuning only part of the LLMs, which may lead to lower performance compared to FedLLMs, where the entire model is updated locally.

In KD-FedLLMs, performance may be affected by indirect optimization. First, the logits derived from a public dataset may not align with the local private dataset, failing to capture the unique features of the local data. This reduces the effectiveness of knowledge transfer between the global and local models. Second, the KD process, which transfers knowledge from the teacher model to the student model, cannot fully replicate the direct parameter updates of FedLLMs, leading to further performance degradation.

\subsection{Communication Overhead}
For FedLLMs, communication costs are determined by a few key factors, including the size of the tunable parameters, such as the LoRA rank and the number of layers to which LoRA is applied. This framework is generally communication-efficient because it only transmits a small set of tunable parameters.

For KD-FedLLMs, the communication overhead is determined by the size of the public dataset and the type of task (size of the logits). 
Firstly, the dimensionality of the logits for each input is determined by the size of the output space, which varies depending on the task type. For classification tasks with $10$ classes, the logit vector has a dimension of $10$, with each element corresponding to the predicted probability of one class. In contrast, for generative tasks, such as language modeling or sequence generation, the logit dimension typically matches the vocabulary size, representing the predicted probability for each token. For example, in a language model with a vocabulary of 50,000 words, the logit vector for each input token would have a dimension of 50,000, each element representing the likelihood of generating a specific token.
Therefore, KD-FedLLMs are most communication-efficient for classification tasks, where logits are relatively small, but become less efficient for generative tasks, where the logit size can be much larger. Additionally, the number of samples in the public dataset also scales the communication overhead, as more data leads to increased logit transmission, further impacting communication efficiency.

In Split-FedLLMs, the communication cost is primarily influenced by the size of the training dataset. This is because, in this framework, clients are required to send activations and gradients to the server for each training sample. As the dataset grows larger, the volume of data exchanged between the client and server increases, leading to higher communication overhead.

\subsection{Client-Side Computation Overhead}
In FedLLMs, clients perform local fine-tuning of the entire LLM, but without the additional tasks of logits generation or distillation, leading to moderate computational overhead.

In KD-FedLLMs, in addition to local fine-tuning, clients are also responsible for generating logits and performing KD, which increases the computational load, resulting in the highest computational workload at the client side.

Split-FedLLMs balance the computational load by partitioning the LLMs between client and server. The client computes the initial layers, while the server handles the deeper layers, resulting in the lowest client-side computational overhead.

\subsection{Overall Comparison and Engineering Guidelines}
The overall comparison of FedLLMs, KD-FedLLMs, and Split-FedLLMs, as summarized in Table \ref{table1}, underscores their respective trade-offs in terms of model accuracy, communication overhead, and client-side computational load.

The strengths and limitations of these frameworks provide valuable engineering insights for selecting and designing the most appropriate frameworks for different fine-tuning scenarios.
\begin{itemize}
    \item FedLLMs prioritize simplicity in the fine-tuning process and optimize model accuracy by directly updating model parameters, making them well-suited for scenarios where high performance is the primary objective. 
    \item KD-FedLLMs enhance communication efficiency by transmitting compact logits, making them advantageous in scenarios where minimizing communication overhead is critical. However, this comes at the cost of a potential reduction in model performance. 
    \item Split-FedLLMs aim to balance the computational workload between client and server, reducing client-side computation, which is particularly beneficial in resource-constrained environments. 
\end{itemize}

\section{Research Opportunities and Future Directions}
Building on the factors impacting each FedLLM framework, this section identifies framework-specific optimization opportunities and explores broader research directions to advance FedLLMs.
\subsection{FedLLM-specifc Research Opportunities}
The design of the tunable parameters largely affects the effectiveness of FedLLMs.
\subsubsection{Optimizing Tunable Parameter Design for Performance and Efficiency Trade-offs}
For FedLLMs, the number of the tunable parameters plays a key role in balancing model performance, communication overhead, and computational load. For instance, in LoRA, smaller ranks of the LoRA matrix reduce communication and computation costs, making them suitable for resource-constrained environments, but may sacrifice model accuracy. Larger ranks enhance model accuracy but incur higher communication and computation resource usage. Thus, one research question is how to design the configuration of tunable parameters for optimization of this trade-off, ensuring high model accuracy while minimizing communication and computational overhead, which is especially critical in bandwidth-limited or resource-constrained FedLLM settings.

\subsubsection{Optimizing Aggregation Strategies for Heterogeneous Clients}
In resource-heterogeneous environments, clients with varying computational capabilities may use different scales of tunable parameters. Clients with limited computational resources may adopt smaller-scale tunable parameters to save resources, while those with higher computational capabilities may utilize larger-scale tunable parameters for enhanced performance.
Thus, it is important to design innovative aggregation strategies that effectively integrate tunable parameters from diverse clients with varying scales. Promising research directions include 1) weighted aggregation methods that account for the parameter scales, 2) adaptive techniques that balance client contributions, and 3) methods such as low-rank approximation to harmonize the scale of tunable parameters during the aggregation process.

\subsubsection{Dynamic Adjustment of Tunable Parameters}
Dynamic adjustment of tunable parameters during training offers an effective way to optimize resource usage. Clients could begin with fewer parameters to reduce initial communication and computation costs, gradually increasing them as the model converges. Additionally, dynamically adjusting the scale of tunable parameters in real-time could optimize resource utilization by adapting to available computational resources and model performance requirements. For instance, clients could increase the parameter size during periods of high resource availability to boost accuracy, while reducing it during low-resource periods to conserve resources.

\subsection{KD-FedLLMs Research Opportunities}
The effectiveness of this framework depends on the quality of the public dataset, while its communication overhead is determined by the sizes of the logits and the public dataset.
\subsubsection{Public Dataset Selection and Alignment} 
 Poor alignment between the public dataset and the clients' private data distributions can result in irrelevant distilled knowledge, negatively impacting global model performance. 
A promising research direction is to improve the alignment between public datasets and clients' private data distributions to enhance the effectiveness of KD-FedLLMs. This can be achieved through adaptive sampling strategies that leverage client feedback and dynamic dataset optimization. For example, clients can provide lightweight information, such as label distribution statistics of the client's dataset. Then, the server can use this information to iteratively refine the public dataset, such as by prioritizing or augmenting samples that better align with the clients’ local data distributions. 

\subsubsection{Logit Dimensions Reduction and Dataset Pruning} 
The communication and computational overheads in KD-FedLLMs increase with the sizes of the logits and the public dataset. Promising techniques to reduce the communication and computational overheads include: 1) dimensionality reduction, such as principal component analysis or low-rank approximation, and filtering methods that retain only the top-$k$ most likely predictions (e.g., tokens with the highest probabilities) for generative tasks, reducing unnecessary data transfer; 2) knowledge compression methods, including transmitting softened label distributions or statistical features like entropy, to further reduce logit size while preserving essential information; 3) dataset pruning, leveraging methods like importance-based sample selection to identify the most representative public data samples, enabling efficient KD with minimal resource usage.

\subsubsection{Advanced Knowledge Processing} 
Upon receiving logits from clients, the server can either select the most relevant ones for distillation or aggregate them to improve the global model. Promising directions for advanced knowledge processing include:
Logit selection techniques, such as importance-based filtering, which prioritizes high-importance logits, or entropy-based filtering, which selects logits with low uncertainty, can help reduce noise and focus on the most informative predictions. Logit aggregation methods, such as weighted aggregation, can assign larger importance to clients with higher data quality or task relevance, while hierarchical clustering can group clients with similar data distributions, enabling more structured and efficient aggregation. 
%A hybrid approach that combines both selection and aggregation offers additional potential, where the server first filters logits based on relevance or confidence and then applies weighted or cluster-based aggregation.

\subsection{Split-FedLLM Research Opportunities}
This approach reduces the computational burden on resource-constrained devices by offloading part of the workload to the server. However, optimizing the partitioning strategy to balance performance, communication efficiency, and computational demands remains crucial, along with the development of communication-efficient algorithms to further reduce communication overhead.

\subsubsection{Dynamic Splitting for Resource-Aware Workload Distribution}
Investigating dynamic partitioning strategies that adaptively adjust the split point between client and server based on heterogeneous resource availability, such as computational power, memory capacity, and network bandwidth. This approach ensures balanced workload distribution, minimizing bottlenecks and improving efficiency in federated learning environments with diverse client capabilities.

\subsubsection{Communication-Efficient Intermediate Activation and Gradient Transfer}
A promising research direction lies in developing techniques to reduce the communication overhead in Split-FedLLMs by optimizing activation and gradient transfer. Adaptive activation compression dynamically prioritizes critical activations, such as leveraging attention-based sparsification to transmit only the most relevant values while discarding less significant ones. Quantized activation transfer applies precision adjustments based on resource constraints, using higher precision for critical layers and lower precision for less significant ones. Additionally, selective data sampling employs importance sampling to identify and transmit the most impactful data points, reducing redundant communication while maintaining model performance.

\subsection{Other General Opportunities}
There are also several general opportunities for advancing FedLLMs and making them more adaptable to real-world applications.
%\subsubsection{Reinforcement Learning for Resource Heterogeneity}
%FL faces the challenge of managing devices with varying computational capacities. Clients in resource-constrained environments struggle with the computational demands of training large models. A promising direction is the incorporation of reinforcement learning (RL) to dynamically adjust model configurations, such as cut-layer placements and communication protocols, based on real-time client resource availability and network conditions. This adaptive approach can optimize client contributions, improving both efficiency and scalability.
%\subsubsection{Generative Models for Data Heterogeneity}
%Data heterogeneity across clients poses a major challenge in federated fine-tuning, especially when clients have non-independent and identically distributed (non-IID) data. %(such as GANs or diffusion models) 
%Generative models could help synthesize synthetic datasets, improving the alignment of client data distributions. These models could generate data that is closer to the global data distribution, enabling better model generalization across clients.

\subsubsection{Continual Learning in Federated Fine-tuning}
Real-world federated learning often involves dynamically changing data distributions and the emergence of new tasks.
A promising direction is to develop frameworks that enable FedLLMs to adapt to new tasks while retaining prior knowledge. Techniques such as memory replay, which stores key representations of past tasks for selective reuse, and regularization-based methods, like Elastic Weight Consolidation (EWC), can help mitigate catastrophic forgetting. Additionally, dynamic model expansion can allocate new parameters or task-specific adapters for incoming tasks while preserving previously learned components. Collaborative knowledge sharing between clients working on similar tasks and task-aware scheduling mechanisms can further enhance adaptability while minimizing resource overhead. 

\subsubsection{Human-in-the-Loop for Real-Time Fine-Tuning}
Incorporating human-in-the-loop (HITL) mechanisms into FedLLMs enables interactive fine-tuning by leveraging real-time user feedback. For instance, applications like personalized chatbots or adaptive recommendation systems can continuously improve by learning from user interactions. Active learning techniques can identify the most informative user data to prioritize during training, reducing communication overhead while maximizing model improvement. By integrating HITL, FedLLMs can achieve enhanced personalization and adaptability, particularly in dynamic environments where user preferences and tasks evolve over time.

\subsubsection{Federated LLMs with Multi-modal Data} 
Many real-world applications require processing multi-modal data, such as text, images, and audio, which demands more complex model architectures and training strategies. Extending FedLLMs to support multi-modal data learning offers a significant research opportunity. Techniques like multi-modal knowledge distillation can transfer information across modalities, enabling efficient learning without sharing raw data. Additionally, adaptive aggregation methods can balance contributions from clients with different data modalities, ensuring effective integration of diverse representations. Developing frameworks to efficiently train and aggregate multi-modal LLMs in federated environments would enable robust and scalable solutions for applications involving diverse data types.

%\subsubsection{Evaluation Metrics for FedLLMs}
%Federated Evaluation Metrics for LLMs
%Challenge: Evaluating federated learning performance is challenging due to varying data distributions and application scenarios.
%Research Opportunity: Define new evaluation metrics for FedLLMs that go beyond traditional metrics like accuracy or BLEU scores. Metrics could include communication efficiency per task, fairness across clients, or generalization across unseen domains. Develop benchmark datasets and evaluation frameworks tailored to federated environments.

\section{Case Study}
To validate the practical value of our analysis, we perform a case study using the GPT-2 model \cite{radford2019language} and the Banking77 dataset \cite{casanueva2020efficient}. This case study further quantitively examines the impact of key factors on the performance of each framework, including LoRA rank $r$, public dataset (PD) size, and number of training samples (TS), offering practical insights into the strengths and limitations of the frameworks under different configurations.
We extract 5002 samples from the training dataset as the public dataset, and the remaining 5001 training data is evenly distributed among three clients. Therefore, each client has a training set size of 1667. In each round, each client trains for one epoch with its training dataset. For our experiments, we implement the LoRA technique with a dropout rate of 0.1 and no bias. The LoRA scaling factor is set to 32, and initialization is performed using a Gaussian distribution. The target module for LoRA is the {attn.c\_attn} layer. The optimization is carried out using the Adam optimizer with a learning rate $\eta = 0.001$. The token padding length is set to 80, and token truncation is enabled. To ensure reproducibility, we use seeds 0, 1, and 42 for the experiments. The results reported are the average of the outcomes obtained from these three seeds.

\begin{figure}[t]
    \centering
    \includegraphics[width=1\linewidth]{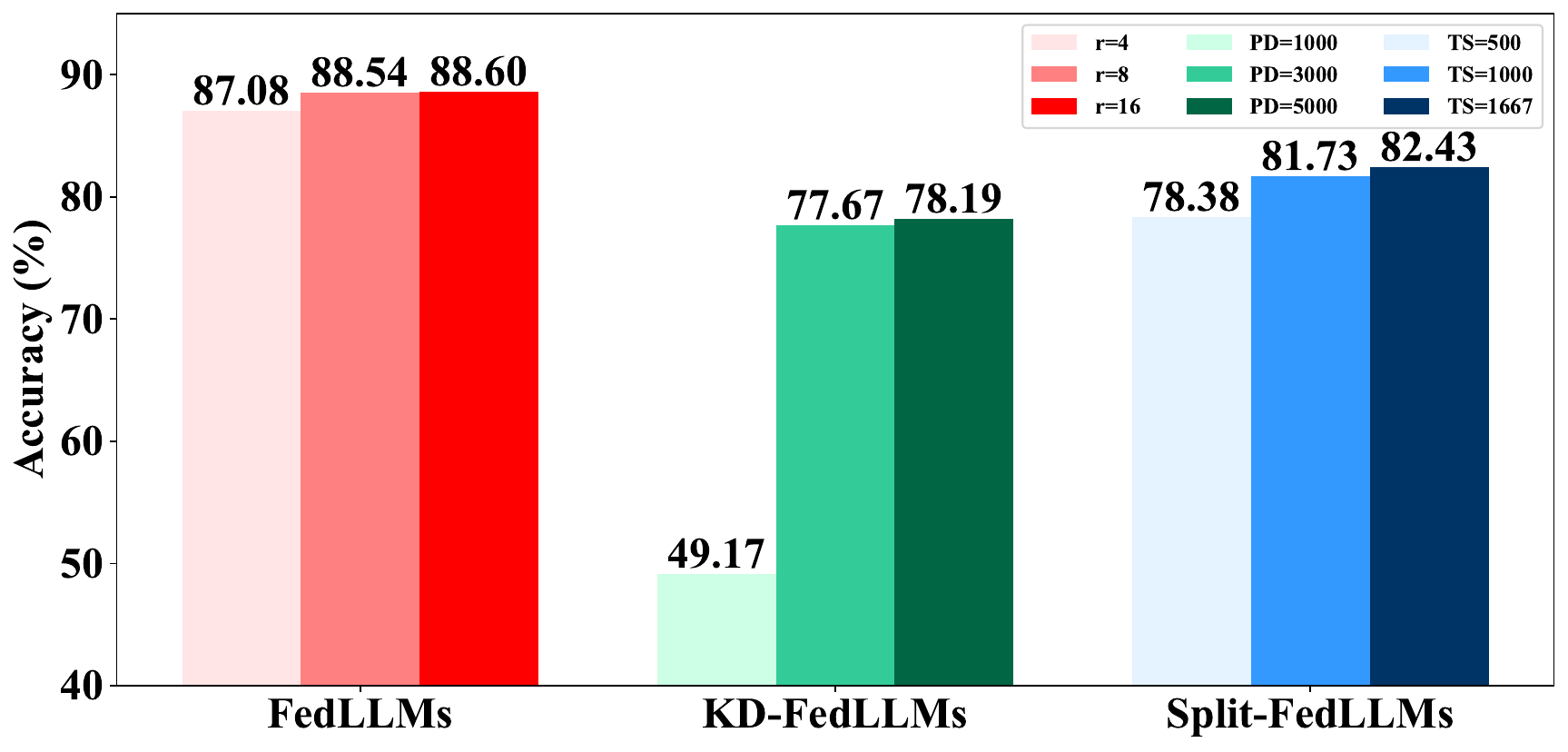}
    \caption{Model accuracy comparison.}
    \label{Server_accuracy}
\end{figure}

Fig.~\ref{Server_accuracy} illustrates the model accuracy after 100 communication rounds for FedLLMs, KD-FedLLMs, and Split-FedLLMs, evaluated under varying configurations of LoRA rank $r$, PD size, and TS size, respectively.
For FedLLMs, we examine the impact of different LoRA ranks, observing that the higher LoRA ranks lead to higher model accuracy. For KD-FedLLMs, we analyze the effect of public dataset size, finding that larger datasets significantly enhance server test accuracy. For Split-FedLLMS, we examine the influence of the number of training samples per communication round, noting that a higher number results in increased model accuracy. Among the three frameworks, FedLLMs achieve the highest model accuracy, outperforming both Split-FedLLMs and KD-FedLLMs. This observation aligns with the qualitative analysis presented in Section \ref{section4}.

\begin{figure}[t]
\includegraphics[width=1.03\linewidth]{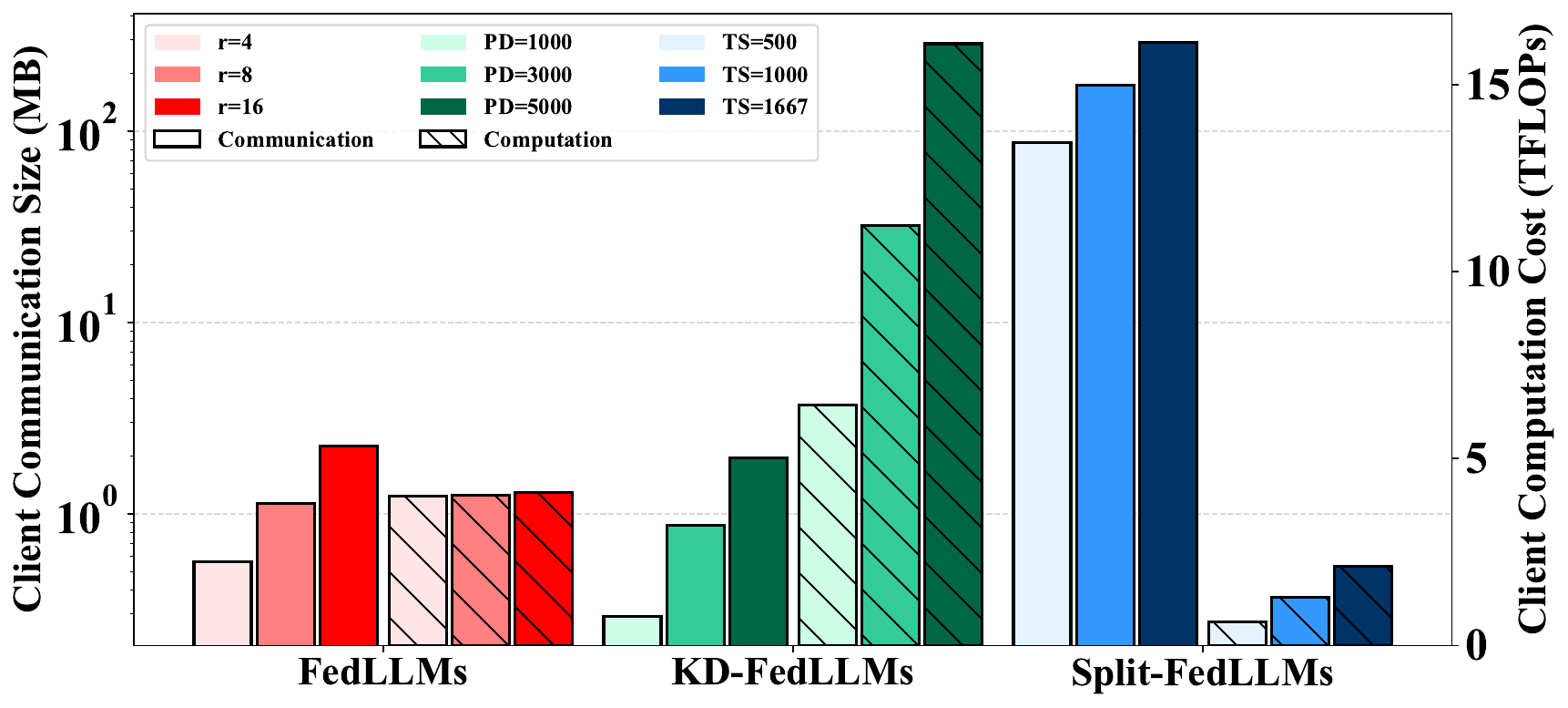}
\caption{Client communication and computation comparison.}
\label{Client_communication_and_computation}
\end{figure}

Fig.~\ref{Client_communication_and_computation} plots the computation and communication overhead for each client per communication round for the three considered federated fine-tuning frameworks. The communication size axis is log-transformed to enhance the visualization of differences across the three frameworks.
In FedLLMs, as the LoRA rank increases, communication costs grow proportionally, whereas the increase in computation costs is relatively small.
In KD-FedLLMs, both communication and computation costs increase proportionally with the size of the public dataset, and KD-FedLLMs exhibit the highest computational cost among the three approaches.
In Split-FedLLMs, communication and computation costs both scale proportionally with the number of training samples per communication round. Split-FedLLMs incur the highest communication overhead among the three frameworks.

\section{Conclusion}
This article provided a comprehensive analysis and comparison of three distributed fine-tuning frameworks for LLMs: FedLLMs, KD-FedLLMs, and Split-FedLLMs. Each framework is based on different knowledge-sharing strategies, including the exchange of model parameters, logits, and activations and gradients, respectively, to facilitate collaborative distributed fine-tuning.
By evaluating their performance in terms of model accuracy, communication overhead, and client-side computational load, we identified the strengths and limitations of each framework, providing valuable insights to guide the selection of the most suitable framework for diverse federated fine-tuning scenarios.
Based on the analysis, several framework-specific optimization opportunities to enhance fine-tuning effectiveness and efficiency were identified. These include adaptive aggregation for heterogeneous clients in FedLLMs, public dataset alignment and pruning in KD-FedLLMs, and dynamic model partitioning with activation compression in Split-FedLLMs.
Additionally, broader research directions were discussed, including continual fine-tuning of FedLLMs, real-time human-in-the-loop feedback, and support for multimodal data.
These insights pave the way for advancing FedLLMs, enabling them to meet the demands of real-world applications.

\bibliographystyle{IEEEtran}
\bibliography{ref}

\end{document}